\documentclass[10pt,twocolumn,letterpaper]{article}
\usepackage{cvpr}
\usepackage{times}
\usepackage{epsfig}
\usepackage{graphicx}
\usepackage{amsmath}
\usepackage{amssymb}
\usepackage{booktabs}
\usepackage[pagebackref=true,breaklinks=true,letterpaper=true,colorlinks,bookmarks=true]{hyperref}
\cvprfinalcopy %
\def\cvprPaperID{****} %

\ifcvprfinal\fi
\begin{document}
\title{ImageNet pre-trained models with batch normalization}
\author{Marcel Simon, Erik Rodner, Joachim Denzler \\
Computer Vision Group \\
Friedrich-Schiller-Universit\"at Jena, Germany\\
{\tt\small \{marcel.simon, erik.rodner, joachim.denzler\}@uni-jena.de}
}
\maketitle
\begin{abstract}
 Convolutional neural networks (CNN) pre-trained on ImageNet are the backbone of most state-of-the-art approaches. 
 In this paper, we present a new set of pre-trained models with popular state-of-the-art architectures for the Caffe framework.
 The first release includes Residual Networks (ResNets) with generation script as well as the batch-normalization-variants of AlexNet and VGG19.
 All models outperform previous models with the same architecture.
 The models and training code are available at \url{http://www.inf-cv.uni-jena.de/Research/CNN+Models.html} and \url{https://github.com/cvjena/cnn-models}.
\end{abstract}
\section{Introduction}
\noindent
The rediscovery of convolutional neural networks (CNN)~\cite{krizhevsky12alexnet} in the past years is a result of both the dramatically increased computational speed and the advent of large scale datasets as part of the big data trend.
The computational speed was mainly boosted by the efficient use of GPUs for common computer vision functions like convolution and matrix multiplication. 
Large scale datasets~\cite{russakovsky15ilsvrc,coco,visualgenome,cocoCaptions,plummer2015flickr30k,cityscapes}, on the other hand, provide the amount of data required for training large scale models with more than a hundred million parameters. 

This combination allowed for huge advances in all fields of computer vision research ranging from traditional tasks like classification~\cite{he15resnet,Simon15:NAC,Simon14:PDD,azizpour14CNNanalysis,Freytag16_CFW,le15deeplysupervised}, object detection~\cite{ren15fasterrcnn,sermanet13overfeat,Gidaris_2015_ICCV}, and segmentation~\cite{long15fcn,Brust15:CPN,crf-nn}, to new ones like image captioning~\cite{johnsonKL15,Rohrbach_2013_ICCV,maoHTCYM15,yuPYBB16,xu15cpation}, visual question answering~\cite{Antol_2015_ICCV,NIPS2015_5641,XiongMS16} and 3D information prediction~\cite{Eigen_2015_ICCV,Wang_2015_CVPR}.
Most of these works are based on models, which are pre-trained on the ImageNet Large Scale Visual Recognition Challenge (ILSVRC) dataset~\cite{russakovsky15ilsvrc}.
The classification task of the last year's ILSVRC contains 1.2 million training images categorized into one thousand categories, which represent a wide variety of everyday objects. 
Pre-training on this dataset proved to be a crucial step for obtaining highly accurate models in most of the tasks mentioned above. 

While computational speed was dramatically increased by the use of GPUs, training a large model like VGG19~\cite{simonyan14vgg} still takes several months on a high-end GPU. 
We hence release a continuously growing set of pre-trained models with popular architectures for the Caffe framework~\cite{jia2014caffe}.
In contrast to most publicly available models for this framework, our release includes the batch normalization~\cite{ioffe15batchnorm} variants of popular networks like AlexNet~\cite{krizhevsky12alexnet} and VGG19~\cite{simonyan14vgg}.
In addition, we provide training code for reproducing the results of residual networks~\cite{he15resnet} in Caffe, which was not provided by the authors of the paper~\cite{resnetModels}.
The release includes all files required for reproducing the model training as well as the log file of the training of the provided model.

\section{Batch normalization for CNNs}
Especially for larger models like VGG19, batch normalization~\cite{ioffe15batchnorm} is crucial for successful training and convergence.
In addition, architectures with batch normalization allow for using much higher learning rates and hence yield in models with better generalization ability. 
In our experiments, we found that higher learning rates show a slower initial convergence speed, but end up at a lower final error rate.
This was the case for both AlexNet and VGG19. 

The advantage of batch normalization is present even for fine-tuning in certain applications.
For example, Amthor~\etal~\cite{Amthor16_IDD} report that their multi-loss architectures only converged reliably if batch normalization was added to the networks. 
However, adding batch normalization afterwards to models trained without batch normalization yields in a severe increase in error rates due to mismatching output statistics.
Instead, fine-tuning with our batch normalization models is directly possible, which allows for easy adaption to new tasks.

\section{Implementation details}
\noindent
We modified AlexNet and VGG19 by adding a batch normalization layer~\cite{ioffe15batchnorm} between each convolutional and activation unit layer as well as between each inner product and activation unit layer. 
We followed the suggestions of \cite{ioffe15batchnorm} and removed the local response normalization and dropout layers.
In addition, we also omitted the mean subtraction during training and replaced it by an batch normalization layer on the input data. 
This results in an adaptively calculated mean in training and relieves users from manually subtracting the mean during feature computation. 
In addition, this approach has the advantage that the mean adapts automatically during fine-tuning and no manual mean calculation and storage is required. 

We train for 64 epochs on the ImageNet Large Scale Visual Recognition Challenge (ILSVRC) 2012 -- 2016 dataset~\cite{russakovsky15ilsvrc}, which contains roughly 1.2 million images and one thousand object categories. 
A batch size of 256 and initial learning rate of 0.05 (AlexNet), 0.01 (VGG19) and 0.1 (ResNet) was used. 
The learning rate follows a linear decay over time.
Due to batch normalization, it is important that the batch size is greater than sixteen to obtain robust statistics estimations in the batch normalization layers. 
In the Caffe framework, this means the batch size in the network definition needs to be sixteen or larger, the solver parameter \texttt{iter\_size} does not compensate a too small batch size in the network definition.
If you want to fine-tune a model but do not have enough GPU memory, you can enable the use of global statistics in training in order to lift this batch size requirement. 
This will disable the statistics estimation in each forward pass and global statistics will be used instead.

All images are resized such that the smaller side has length 256 pixel and the aspect ratio is preserved. 
During training, we randomly crop a $224\times 224$ (ResNet, VGG19) or $227\times 227$ (AlexNet) pixel square patch and feed it into the network.
During validation, a single centered crop is used. 
We did not use any kind of color, scale or aspect ratio augmentation.

During training of residual networks, we also observe a sudden divergence at random time points in training as explained by Szegedy~\etal~\cite{szegedy16inception4}.
In this case, we restarted the training using the last snapshot.
Due to a different random seed, the order of the images is different and hence the training does not diverge at this time point anymore. 

Please note, that the final models are not cherry picked based on the validation error.
We provide the final model after the full training is completed. 
We did not intervene with training and especially did not manually changed the learning rate, as usually done if the step policy is used for the learning rate.

\section{Results}
\noindent
The top-1 and top-5-error of the trained models are shown in Table~\ref{tab:model-acc}.
\begin{table}
\caption{\textbf{Single-crop} top-1 and top-5 error of our models on the validation set of ILSVRC 2012. 
\label{tab:model-acc}}
 \centering
 \begin{tabular}{lcc|cc}
  \toprule
  Model &\multicolumn{2}{c|}{Top-1 error}&\multicolumn{2}{c}{Top-5 error}\\
        &  Ours & Original 		 	&  Ours & Original \\
  \midrule
  AlexNet  &\textbf{39.9\%}	&42.6\% 	&\textbf{18.1\%}& 19.6\% 	\\
  VGG19    &\textbf{26.9\%}	&28.7\%		&\textbf{8.8\%}	& 9.9\% 	\\
  ResNet-10&\textbf{36.1\%}	& --		&\textbf{14.8\%}& --    	\\
  ResNet-50&\textbf{24.6\%}	&24.7\% 	&\textbf{7.6\%}	& 7.8\%		\\
  \bottomrule
 \end{tabular}
\end{table}
As observed in previous works~\cite{ioffe15batchnorm}, the error rates benefit from the added batch normalization layers. 
All provided models slightly improve the error rate achieved by previously trained models~\cite{matconvnetPretrained}.
In case of AlexNet, for example, we even observe an error decrease of over 2.6\%.

\begin{figure}
 \centering
 \includegraphics[width=\linewidth]{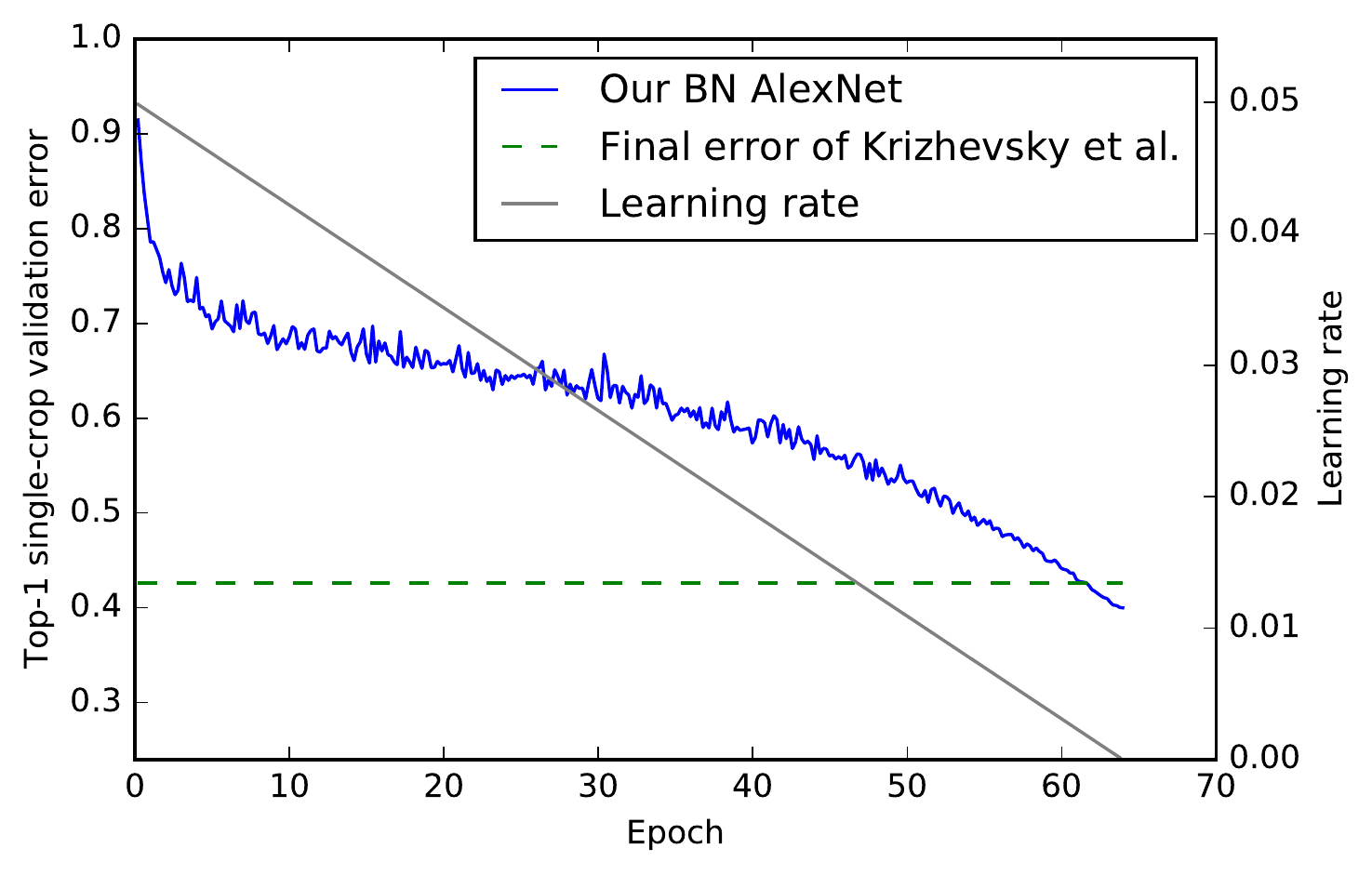}
 \caption{\textbf{Single crop} top-1 error of AlexNet on the validation set of ILSVRC 2012 with respect to the training time.
 We used a linear learning rate decay as shown by the gray curve, which explains the steep decrease in error towards the end of the training.
 \label{fig:alexnet-training}}
\end{figure}

In addition to the final results, we also visualize the single-crop top-1 error on the validation set during the training of AlexNet in Fig.~\ref{fig:alexnet-training}.
As shown in the figure, the error decreases consistently and fairly quickly during training. 
Since we use linear learning rate decay, there is a steep error decrease towards the end of the training. 
While it might look like the error could decrease even further, this is not true. 
The reason is that the learning rate approaches 0 during the end of the training. 
Even if the learning rate is kept constant, no improvement can be observed.
This is supported by several experiments we performed. 

\section{Conclusions}
This paper presents a new set of pre-trained models for the ImageNet dataset using the Caffe framework.
We focus on the batch-normalization-variants of AlexNet and VGG19 as well as residual networks.
All models outperform previous pre-trained models.
In particular, we were able to reproduce the ImageNet results of residual networks.
All models, log files and training code are available at \url{http://www.inf-cv.uni-jena.de/Research/CNN+Models.html} and \url{https://github.com/cvjena/cnn-models}.

\section{Acknowledgments}
\noindent
The authors thank Nvidia for GPU hardware donations. 
Part of this research was supported by grant RO 5093/1-1 of the German Research Foundation (DFG)

{\small
\bibliographystyle{ieee}
\bibliography{paper}
}

\end{document}